\documentclass{article}
\usepackage{times}%
\usepackage{graphicx} 
\usepackage{caption}
\usepackage{subcaption}

\usepackage{natbib}

\usepackage{algorithmic}

\usepackage{hyperref}

\usepackage[accepted]{icml2017}

\icmltitlerunning{Towards K-means-friendly Spaces: Simultaneous Deep Learning and Clustering}
\usepackage{calc,amsfonts,amssymb,amsmath,bm,url,color,theorem}
\usepackage{graphicx}
\usepackage{psfrag,float}

\usepackage{multirow}
\usepackage{epstopdf}
\usepackage{color}
\usepackage[dvipsnames]{xcolor}

\usepackage{epstopdf,nicefrac}
\usepackage{algorithm}
\usepackage{algorithmic}

\DeclareMathOperator*{\argmin}{arg\,min}

\newcommand{\R}{\mathbb{R}}
\newcommand{\W}{\mathcal{W}}

\definecolor{orange}{RGB}{255,107,0}

\begin{document}

\twocolumn[
\icmltitle{Towards K-means-friendly Spaces: Simultaneous Deep \\
 Learning and Clustering}

\begin{icmlauthorlist}
\icmlauthor{Bo Yang}{umn}
\icmlauthor{Xiao Fu}{umn}
\icmlauthor{Nicholas D. Sidiropoulos}{umn}
\icmlauthor{Mingyi Hong}{isu}
\end{icmlauthorlist}

\icmlaffiliation{umn}{Department of Electrical and Computer Engineering,
	University of Minnesota, Minneapolis MN 55455, USA.}
\icmlaffiliation{isu}{Department of Industrial and Manufacturing Systems Engineering, Iowa State University, Ames,  IA 50011, USA}

\icmlcorrespondingauthor{Bo Yang}{yang4173@umn.edu}
\icmlcorrespondingauthor{Xiao Fu}{xfu@umn.edu}
\icmlcorrespondingauthor{Nicholas D. Sidiropoulos}{nikos@ece.um.edu}
\icmlcorrespondingauthor{Mingyi Hong}{mingyi@iastate.edu}
\icmlkeywords{boring formatting information, machine learning, ICML}



\vskip 0.3in
]

\printAffiliationsAndNotice{}  

%

\vspace{-.35cm}
\begin{abstract} 
\vspace{-.3cm}
Most learning approaches treat dimensionality reduction (DR) and clustering separately (i.e., sequentially), but recent research has shown that optimizing the two tasks jointly can substantially improve the performance of both. The premise behind the latter genre is that the data samples are obtained via linear transformation of latent representations that are easy to cluster; but in practice, the transformation from the latent space to the data can be more complicated. In this work, we assume that this transformation is an unknown and possibly \emph{nonlinear} function.
To recover the `clustering-friendly' latent representations and to better cluster the data, we propose a joint DR and K-means clustering approach in which DR is accomplished via learning a deep neural network (DNN). 
The motivation is to keep the advantages of jointly optimizing the two tasks, while exploiting the deep neural network's ability to approximate any nonlinear function. 
This way, the proposed approach can work well for a broad class of generative models. 
Towards this end, we carefully design the DNN structure and the associated joint optimization criterion, and propose an effective and scalable algorithm to handle the formulated optimization problem. Experiments using different real datasets are employed to showcase the effectiveness of the proposed approach.

\end{abstract}

\section{Introduction}
Clustering is one of the most fundamental tasks in data mining and machine learning, with an endless list of applications. 
It is also a notoriously hard task, whose outcome is affected by a number of factors -- including data acquisition and representation, use of preprocessing such as dimensionality reduction (DR), the choice of clustering criterion and optimization algorithm, and initialization  \cite{ertoz2003finding, banerjee2005clustering}.
Since its introduction in 1957 by Lloyd (published much later in 1982 \cite{lloyd1982least}), K-means has been extensively used either alone or together with suitable preprocessing, due to its simplicity and effectiveness. 
K-means is suitable for clustering data samples that are evenly spread around some centroids (cf. the first subfigure in Fig.~\ref{fig: synthetic data}), but many real-life datasets do not exhibit this `K-means-friendly' structure.
Much effort has been spent on mapping high-dimensional data to a certain space that is suitable for performing K-means.
Various techniques, including principal component analysis (PCA), canonical correlation analysis (CCA), nonnegative matrix factorization (NMF) and sparse coding (dictionary learning), were adopted for this purpose. 
In addition to these linear DR operators (e.g., a projection matrix), nonlinear DR techniques such as those used in spectral clustering \cite{ng2002spectral} and sparse subspace clustering  \cite{elhamifar2013sparse, you2016scalable} have also been considered.


In recent years, motivated by the success of deep neural networks (DNNs) in {\em supervised} learning, {\em unsupervised} deep learning approaches are now widely used for DR prior to clustering. For example, the stacked autoencoder (SAE) \cite{vincent2010stacked}, deep CCA (DCCA) \cite{andrew2013deep}, and sparse autoencoder \cite{ng2011sparse} take insights from PCA, CCA, and sparse coding, respectively, and make use of DNNs to learn nonlinear mappings from the data domain to
low-dimensional latent spaces. These approaches treat their DNNs as a preprocessing stage that is separately designed from the subsequent clustering stage. The hope is that the latent representations of the data learned by these DNNs will be naturally suitable for clustering. However, since no clustering-promoting objective is explicitly incorporated in the learning process, the learned DNNs do not necessarily output reduced-dimension data that are suitable for clustering -- as will be seen in our experiments.

In \cite{de1994k,patel2013latent,yang2016learning}, joint DR and clustering was considered.
The rationale behind this line of work is that if there exists {\em some} latent space where the entities nicely fall into clusters, then it is natural to seek a DR transformation that reveals such structure, i.e., which yields a low K-means clustering cost. 
This motivates using the K-means cost in latent space as a prior that helps choose the right DR, and pushes DR towards producing K-means-friendly representations. By performing joint DR and K-means clustering, impressive clustering results have been observed in \cite{yang2016learning}. The limitation of these works is that the observable data is assumed to be generated from the latent clustering-friendly space via simple linear transformation. While simple linear transformation works well in many cases, there are other cases where the generative process is more complex, involving a nonlinear mapping.

\noindent
{\bf Contributions} In this work, we propose a joint DR and K-means clustering framework, where the DR part is implemented through learning a DNN, rather than a linear model. Unlike previous attempts that utilize this joint DNN and clustering idea, we made customized design for this \emph{unsupervised} task.
Although implementing this idea is highly non-trivial (much more challenging than \cite{de1994k,patel2013latent,yang2016learning} where the DR part only needs to learn a linear model), 
our objective is well-motivated: by better modeling the data transformation process with a more general model, a much more K-means-friendly latent space can be learned -- as we will demonstrate.  A sneak peek of the kind of performance that can be expected using our proposed method can be seen in Fig.~\ref{fig: synthetic data}, where we generate four clusters of 2-D data which are well separated in the 2-D Euclidean space and then transform them to a 100-D space using a complex non-linear mapping [cf. \eqref{eq: sigsig}] which destroys the cluster structure. 
One can see that the proposed algorithm outputs reduced-dimension data that are most suitable for applying K-means. Our specific contributions are as follows:

\begin{figure}
	\centering
	\includegraphics[width=0.95\linewidth]{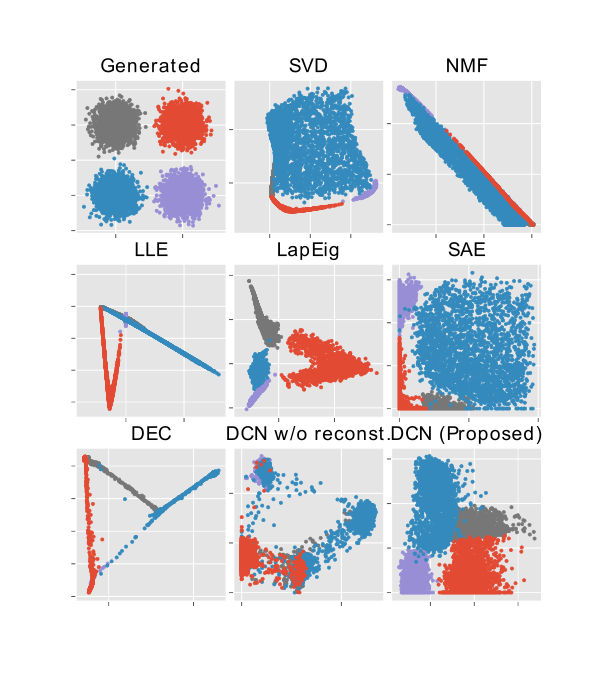}
	\vspace{-1cm}
	\caption{The learned 2-D reduced-dimension data by different methods. The observable data is in the 100-D space and is generated from 2-D data (cf. the first subfigure) through the nonlinear transformation in \eqref{eq: sigsig}. The true cluster labels are indicated using different colors.}
	\label{fig: synthetic data}	
	\vspace{-.5cm}
\end{figure}

\noindent
$\bullet$ {\bf Optimization Criterion Design}:
We propose an optimization criterion for joint DNN-based DR and K-means clustering.
The criterion is a combination of three parts, namely, dimensionality reduction, data reconstruction, and cluster structure-promoting regularization.
We deliberately include the reconstruction part and implement it using a decoding network, which is crucial for avoiding trivial solutions.
The criterion is also flexible -- it can be extended to incorporate different DNN structures (e.g. convolutional neural networks \cite{lecun1998gradient, krizhevsky2012imagenet}) and clustering criteria, e.g., subspace clustering.

\noindent
$\bullet$ {\bf Effective and Scalable Optimization Procedure}:
The formulated optimization problem is very challenging to handle, since it involves layers of nonlinear activation functions and integer constraints that are induced by the K-means part.
We propose a judiciously designed solution package, including empirically effective initialization and a novel alternating stochastic gradient algorithm. The algorithmic structure is simple, enables online implementation, and is very scalable.

\noindent
$\bullet$ {\bf Comprehensive Experiments and Validation}:
We provide a set of synthetic-data experiments and validate the method on different real datasets including various document and image copora. Evidently visible improvement from the respective state-of-art is observed for all the datasets that we experimented with.

\noindent
$\bullet$ {\bf Reproducibility}: The code for the experiments is available at \url{https://github.com/boyangumn/DCN}.
	

\vspace{-.25cm}
\section{Background and Related Works}\label{ch: background}
Given a set of data samples $\{{\bm x}_i\}_{i=1,\ldots,N}$ where ${\bm x}_i\in\mathbb{R}^M$, the task of clustering is to group the $N$ data samples into $K$ categories.  Arguably, K-means \cite{lloyd1982least} is the most widely adopted algorithm. 
K-means approaches this task by optimizing the following
cost function:
\begin{align}\label{eq:Kmeans}
\min_{{\bm M}\in\mathbb{R}^{{\bm M}\times K}, \{{\bm s}_i\in\mathbb{R}^K\}} &~~  \sum_{i=1}^{N}\left\|{\bm x}_i - {\bm M}{\bm s}_i \right\|_2^2 \\	
\text{s.t.} &~~ s_{j,i} \in \{0,1\}, ~ {\bm 1}^T{\bm s}_i = 1 ~~\forall i, j,  \nonumber
\end{align}
where ${\bm s}_i$ is the assignment vector of data point $i$ which has only one non-zero element,
$s_{j,i}$ denotes the $j$th element of ${\bm s}_i$, and the $k$th column of ${\bm M}$, i.e., ${\bm m}_k$, denotes the centroid of the $k$th cluster.

K-means works well when the data samples are evenly scattered around their centroids in the feature space; we consider datasets which have this structure as being `K-means-friendly' (cf. top-left subfigure of Fig.~\ref{fig: synthetic data}).  
However, high-dimensional data are in general not very K-means-friendly. In practice, using a DR pre-processing, e.g., PCA or NMF \cite{xu2003document, cai2011locally}, to reduce the dimension of ${\bm x}_i$ to a much lower dimensional space and then apply K-means usually gives better results.
In addition to the above classic DR methods that essentially learn a linear generative model from the latent space to the data domain, nonlinear DR approaches such as those used in spectral clustering \cite{ng2002spectral, von2007tutorial} and DNN-based DR \cite{hinton2006reducing,schroff2015facenet, hershey2016deep}  are also widely used as pre-processing before K-means or other clustering algorithms, see also \cite{vincent2010stacked,bruna2013invariant}.

Instead of using DR as a pre-processing, joint DR and clustering was also considered in the literature \cite{de1994k,patel2013latent, yang2016learning}. This line of work can be summarized as follows. Consider the generative model where a data sample is generated by ${\bm x}_i={\bm W}{\bm h}_i$, where ${\bm W}\in\mathbb{R}^{M\times R}$ and ${\bm h}_i\in\mathbb{R}^{R}$, where $R\ll M$.
Assume that the data clusters are well-separated in latent domain (i.e., where ${\bm h}_i$ lives) but distorted by the transformation introduced by ${\bm W}$. Reference \cite{yang2016learning} formulated the joint optimization problem as follows:
\begin{align}\label{eq:JNKM}
\min_{{\bm M}, \{{\bm s}_i\},{\bm W},{\bm H}} &~~ \left\|{\bm X}-{\bm W}{\bm H}\right\|_F^2+ \lambda\sum_{i=1}^{N}\left\|{\bm h}_i - {\bm M}{\bm s}_i \right\|_2^2 \nonumber\\
&~~~~~ + r_1({\bm H})+r_2({\bm W}) \\	
\text{s.t.} &~~ s_{j,i} \in \{0,1\}, ~ {\bm 1}^T{\bm s}_i = 1 ~~\forall i, j,  \nonumber
\end{align}
where ${\bm X}=[{\bm x}_1,\ldots,{\bm x}_N]$, ${\bm H}=[{\bm h}_1,\ldots,{\bm h}_N]$,
and $\lambda\geq 0$ is a parameter for balancing data fidelity and the latent cluster structure.
In \eqref{eq:JNKM}, the first term performs DR and the second term performs latent clustering.
The terms $r_1(\cdot)$ and $r_2(\cdot)$ are regularizations (e.g., nonnegativity or sparsity)
to prevent trivial solutions, e.g., ${\bm H}\rightarrow{\bm 0}\in\mathbb{R}^{R\times N}$; see details in \cite{yang2016learning}.

The data model ${\bm X} \approx{\bm W}{\bm H}$ in the above line of work may be oversimplified: The data generating process can be much more complex than this linear transform. Therefore, it is well justified to seek powerful non-linear transforms, e.g. DNNs, to model this data generating process, while at the same time make use of the joint DR and clustering idea. Two recent works, \cite{xie2015unsupervised} and \cite{JULE}, made such attempts. 

\begin{figure}
\centering
   \includegraphics[width=0.75\linewidth]{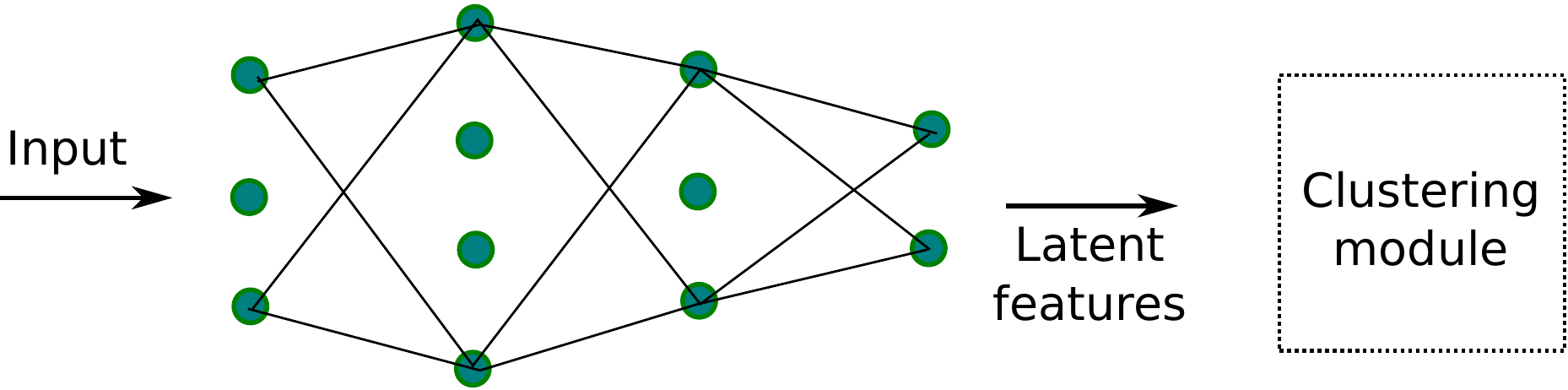}
\caption{A problematic \emph{joint} deep clustering structure. To avoid clutter,  some links are omitted.}
\vspace{-.5cm}
\label{fig:DEC}
\end{figure}

The idea of \cite{xie2015unsupervised} and \cite{JULE} is to connect a clustering module to the output layer of a DNN, and jointly learn DNN parameters and clusters. Specifically, the approaches look into an optimization problem of the following form
\begin{align}\label{eq: plain}
\min_{\mathcal{W}, \Theta} &~~ \widehat{L} = \sum_{i=1}^{N}q({\bm f}({\bm x}_i;{\cal W}); \Theta), 
\end{align}
where ${\bm f}({\bm x}_i;{\cal W})$ is the network output given data sample ${\bm x}_i$, ${\cal W}$ collects the network parameters, and $\Theta$ denotes parameters of some clustering model. For instance, $\Theta$ stands for the centroids ${\bm M}$ and assignments $ \{{\bm s}_i\}$ if the K-means clustering formulation \eqref{eq:Kmeans} is adopted. The $q(\cdot)$ in \eqref{eq: plain} denotes some clustering loss, e.g., the Kullback-Leibler (KL) divergence loss in \cite{xie2015unsupervised} and agglomerative clustering loss in \cite{JULE}. An illustration of this kind of approaches is shown in Fig.~\ref{fig:DEC}. This idea seems reasonable, but is problematic. 
A \emph{global optimal} solution to Problem~\eqref{eq: plain} is $f({\bm x}_i;{\cal W})={\bm 0}$ and the optimal objective value $\widehat{L}=0$ can always be achieved.
Another type of trivial solutions are simply mapping \emph{arbitrary} data samples to tight clusters, which will lead to a small value of $\widehat{L}$ -- but this could be far from being desired since there is no provision for respecting the data samples ${\bm x}_i$'s; see the bottom-middle subfigure in Fig.~\ref{fig: synthetic data} [Deep Clustering Network (DCN) w/o reconstruction] and the bottom-left subfigure  in Fig.~\ref{fig: synthetic data} [DEC].
This issue also exists in \cite{JULE}.

\vspace{-.25cm}
\section{Proposed Formulation}
We are motivated to model the relationship between the observable data ${\bm x}_i$ and its clustering-friendly latent representation ${\bm h}_i$ using a nonlinear mapping, i.e.,
\[  {\bm h}_i = {\bm f}({\bm x}_i;{\cal W}),\quad {\bm f}(\cdot;{\cal W}):~\mathbb{R}^M \rightarrow \mathbb{R}^R, \]
where ${\bm f}(\cdot;{\cal W})$ denotes the mapping function and ${\cal W}$ denote the set of parameters.
In this work, we propose to employ a DNN as our mapping function, since DNNs have the ability of approximating any continuous mapping using a reasonable number of parameters \cite{hornik1989multilayer}.

We want to learn the DNN and perform clustering \emph{simultaneously}. The critical question here is how to avoid trivial solutions in this \emph{unsupervised} task. In fact, this can be resolved by taking insights from \eqref{eq:JNKM}. 
The key to prevent trivial solution in the linear DR case lies in the reconstruction part, i.e., the term $\left\|{\bm X}-{\bm W}{\bm H}\right\|_F^2$ in \eqref{eq:JNKM}. This term ensures that the learned ${\bm h}_i$'s can (approximately) reconstruct the ${\bm x}_i$'s using the basis ${\bm W}$.
This motivates incorporating a reconstruction term in the joint DNN-based DR and K-means.
In the realm of unsupervised DNN, there are several well-developed approaches for reconstruction -- e.g., the stacked autoencoder (SAE) is a popular choice for serving this purpose.
To prevent trivial low-dimensional representations such as all-zero vectors,
SAE uses a decoding network ${\bm g}(\cdot;{\cal Z})$ to map the ${\bm h}_i$'s back to the data domain and requires that
${\bm g}({\bm h}_i;{\cal Z})$ and ${\bm x}_i$ match each other well under some metric, e.g., mutual information or least squares-based measures.

By the above reasoning, we come up with the following cost function:
\begin{align}\label{eq: formulation}
	\min_{\W,{\cal Z}, \atop {\bm M}, \{{\bm s}_i\}} &~ \sum_{i=1}^{N}\left ( \ell \left({\bm g}({\bm f}({\bm x}_i)), {\bm x}_i\right)  + \frac{\lambda}{2} \left\|{\bm f}({\bm x}_i) - {\bm M}{\bm s}_i \right\|_2^2 \right)\\	
	\text{s.t.} &~~ s_{j,i} \in \{0,1\}, ~ {\bm 1}^Ts_i = 1 ~~\forall i, j, \nonumber
\end{align}
where we have simplified the notation ${\bm f}({\bm x}_i;{\cal W})$ and ${\bm g}({\bm h}_i;{\cal Z})$ to ${\bm f}({\bm x}_i)$ and ${\bm g}({\bm h}_i)$, respectively, for conciseness.
The function $\ell(\cdot): \mathbb{R}^{M}\rightarrow \mathbb{R}$ is a certain loss function that measures the reconstruction error. In this work, we adopt the least-squares loss $\ell({\bm x}, {\bm y}) = \left\|{\bm x} - {\bm y}\right\|_2^2$;
other choices such as $\ell_1$-norm based fitting and the KL divergence can also be considered.
$\lambda\geq 0$ is a regularization parameter which balances the reconstruction error versus finding K-means-friendly latent representations.

%

\begin{figure}
\centering
\includegraphics[width=\linewidth]{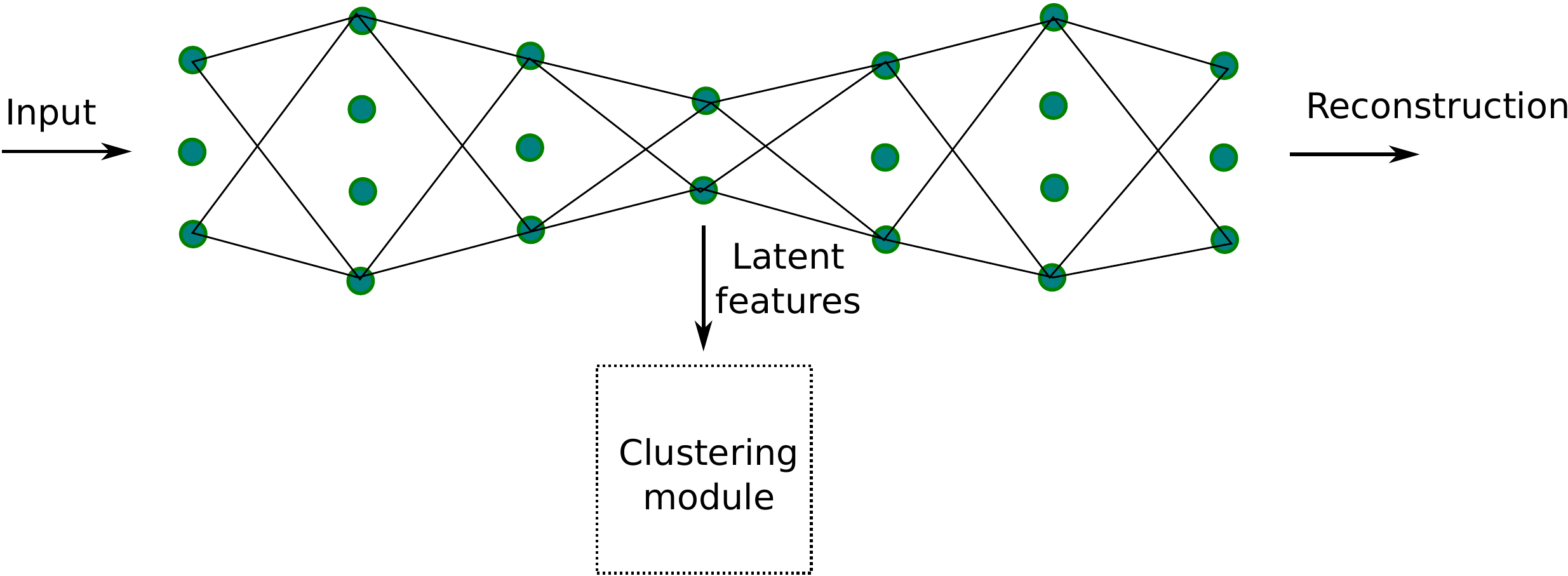}
\caption{Proposed deep clustering network (DCN).}
\vspace{-.5cm}
\label{fig:DNNexample}
\end{figure}

Fig.~\ref{fig:DNNexample} presents the network structure corresponding to the formulation in \eqref{eq: formulation}. Compare to the network in Fig.~\ref{fig:DEC}, our latent features are also responsible for reconstructing the input, preventing all the aforementioned trivial solutions.
On the left-hand side of the `bottleneck' layer are the so-called encoding or forward layers that transform raw data to a low-dimensional space.
On the right-hand side are the `decoding' layers that try to reconstruct the data from the latent space.
The K-means task is performed at the bottleneck layer.
The forward network, the decoding network, and the K-means cost are optimized simultaneously.
In our experiments, the structure of the decoding networks is a `mirrored version' of the encoding network, and for both the encoding and decoding networks, we use the \emph{rectified linear unit} (ReLU) activation-based neurons \cite{nair2010rectified}. Since our objective is to perform DNN-driven K-means clustering, we will refer to the network in Fig.~\ref{fig:DNNexample} as the Deep Clustering Network (DCN) in the sequel.

We should remark that the proposed optimization criterion in \eqref{eq: formulation} and the network in Fig.~\ref{fig:DNNexample} are very flexible: Other types of networks, e.g., deep convolutional neural networks \cite{lecun1998gradient, krizhevsky2012imagenet}, can be used. For the clustering part, other clustering criteria, e.g., K-subspace and soft K-means \cite{law2005model, banerjee2005clustering}, are also viable options. Nevertheless, we will concentrate on the proposed DCN in the sequel, as our interest is to provide a proof-of-concept rather than exhausting the possibilities of combinations.

\vspace{-.25cm}
\section{Optimization Procedure}
Optimizing \eqref{eq: formulation} is highly non-trivial since both the cost function and the constraints are non-convex.
In addition, there are scalability issues that need to be taken into account.
In this section, we propose a pragmatic optimization procedure including an empirically effective initialization method and an alternating optimization based algorithm for handling \eqref{eq: formulation}.

\vspace{-.25cm}
\subsection{Initialization via Layer-wise Pre-Training}
\label{sub:layer_wise_pre_training}
For dealing with hard non-convex optimization problems like that in \eqref{eq: formulation}, initialization is usually crucial.
To initialize the parameters of the network, i.e., $({\cal W},{\cal Z})$, we use
the layer-wise pre-training method as in \cite{bengio2007greedy} for training autoencoders. This pre-training technique may be avoided in large-scale supervised learning tasks. For the proposed DCN which is completely unsupervised, however, we find that the layer-wise pre-training procedure is important no matter the size of the dataset. We refer the readers to \cite{bengio2007greedy} for an introduction of layer-wise pre-training.
After pre-training, we perform K-means to the outputs of the bottleneck layer to obtain initial values of ${\bm M}$ and $\{{\bm s}_i\}$.

\vspace{-.25cm}
\subsection{Alternating Stochastic Optimization} 
\label{sub:fine_tuning}
Even with a good initialization, handling Problem~\eqref{eq: formulation} is still very challenging.
The commonly used stochastic gradient descent (SGD) algorithm cannot be directly applied to
jointly optimize ${\cal W},{\cal Z},{\bm M}$ and $\{{\bm s}_i\}$ because the block variable $\{{\bm s}_i\}$ is constrained on a discrete set.
Our idea is to combine the insights of alternating optimization and SGD.
Specifically, we propose to optimize the subproblems with respect to (w.r.t.) one of ${\bm M}$, $\{{\bm s}_i\}$ and $({\cal W},{\cal Z})$ while keeping the other two sets of variables fixed.

\subsubsection{Update network parameters}
For fixed $({\bm M},\{{\bm s}_i\})$, the subproblem w.r.t. $({\cal W},{\cal Z})$ is similar to training an SAE -- but with an additional penalty term on the clustering performance.
We can take advantage of the mature tools for training DNNs, e.g., back-propagation based SGD and its variants.
To implement SGD for updating the network parameters, we look at the problem w.r.t. the incoming data ${\bm x}_i$:
\begin{align}\label{eq: sub_network}
\min_{{\cal W},{\cal Z}} &~L^i=\ell\left({\bm g}({\bm f}({\bm x}_i)), {\bm x}_i\right)  + \frac{\lambda}{2} \left\|{\bm f}({\bm x}_i) - {\bm M}{\bm s}_i \right\|_2^2.
\end{align}
The gradient of the above function over the network parameters is
easily computable, i.e.,
$	\triangledown_{{\cal X}} L^{i}  = \frac{\partial \ell\left({\bm g}({\bm f}({\bm x}_i)), {\bm x}_i\right) }{\partial {\cal X}}+ \lambda\frac{\partial {\bm f}({\bm x}_i)}{\partial \mathcal{X}}({\bm f}({\bm x}_i) - {\bm M}{\bm s}_i)$,
where ${\cal X}=({\cal W},{\cal Z})$ is a collection of the network parameters
and the gradients $\frac{\partial \ell}{\partial {\cal X}}$ and $\frac{\partial {\bm f}({\bm x}_i)}{\partial \mathcal{X}}$ can be calculated by back-propagation \cite{rumelhart1988learning} (strictly speaking, what we calculate here is the \emph{subgradient} w.r.t. ${\cal X}$ since the ReLU function is non-differentible at zero).
Then, the network parameters are updated by
\begin{align}\label{eq: net_update}
	\mathcal{X} \leftarrow \mathcal{X} - \alpha \triangledown_{\cal X} L^{i},
\end{align}
where $\alpha>0$ is a diminishing learning rate.

\subsubsection{Update clustering parameters} 
For fixed network parameters and ${\bm M}$, the assignment vector of the current sample, i.e., ${\bm s}_i$, can be naturally updated in an online fashion.
Specifically, we update ${\bm s}_i$ as follows:
\begin{align}\label{eq: S_update}
	s_{j,i} \leftarrow \begin{cases}
	1, \quad \text{if } j =  \underset{{k=\{1,\ldots,K\}} }{\argmin}\left\|{\bm f}({\bm x}_i) - {\bm m}_k \right\|_2, \\
	0, \quad \text{otherwise.}
	\end{cases}
\end{align}

When fixing $\{{\bm s}_i\}$ and ${\cal X}$, the update of ${\bm M}$ is simple and may be done in a variety of ways.
For example, one can simply use ${\bm m}_k=(1/|{\cal C}_k^i|)\sum_{i\in{\cal C}_k^i}{\bm f}({\bm x}_i)$, where ${\cal C}_k^i$ is the recorded index set of samples assigned to cluster $k$ from the first sample to the current sample $i$.
Although the above update is intuitive, it could be problematic for online algorithms, since the already appeared historical data (i.e., ${\bm x}_1,\ldots,{\bm x}_i$) might not be representative enough to model the global cluster structure and the initial ${\bm s}_i$'s might be far away from being correct.
Therefore, simply averaging the current assigned samples may cause numerical problems.
Instead of doing the above, we employ the idea in \cite{sculley2010web}
to adaptively change the learning rate of updating ${\bm m}_1,\ldots,{\bm m}_K$.
The intuition is simple: assume that the clusters are roughly balanced in terms of the number of data samples they contain. Then, after updating ${\bm M}$ for a number of samples, one should update the centroids of the clusters that already have many assigned members more gracefully while updating others more aggressively, to keep balance.
To implement this, let ${c}_k^i$ be the count of the number of times the algorithm assigned a sample to cluster $k$ before handling the incoming sample ${\bm x}_i$, and update ${\bm m}_k$ by a simple gradient step:
\begin{align}\label{eq: M_update}	
	{\bm m}_k & \leftarrow {\bm m}_k - (\nicefrac{1}{{c}_k^i})\left({\bm m}_k - {\bm f}({\bm x}_i)\right){s}_{k, i},
\end{align}
where the gradient step size $1/c_k^i$ controls the learning rate. The above update of ${\bm M}$ can also be viewed as an SGD step, thereby resulting in an overall alternating block SGD procedure that is summarized in Algorithm~\ref{algo:proposed}. Note that an epoch corresponds to a pass of all data samples through the network.

\begin{algorithm}
	\begin{algorithmic}[1]
		\STATE {Initialization}
		\COMMENT{Perform $T$ epochs over the data}
		\FOR {$t = 1:T$}
		\STATE {Update network parameters by \eqref{eq: net_update}}
		\STATE {Update assignment by \eqref{eq: S_update}}
		\STATE {Update centroids by \eqref{eq: M_update}}
		\ENDFOR
	\end{algorithmic}
	\caption{Alternating SGD}\label{algo:proposed}
\end{algorithm}

Algorithm~\ref{algo:proposed} has many favorable properties.
First, it can be implemented in a completely online fashion, and thus is very scalable.
Second, many known tricks for enhancing performance of DNN training can be directly used. In fact, we have used a mini-batch version of SGD and batch-normalization \cite{ioffe2015batch} in our experiments, which indeed help improve performance.


\vspace{-.25cm}
\section{Experiments}
In this section, we use synthetic and real-world data to showcase the effectiveness of DCN.
We implement DCN using the deep learning toolbox Theano \cite{theano}. 

%

\vspace{-.25cm}
\subsection{Synthetic-Data Demonstration} \label{refsubsec}
Our settings are as follows: Assume that the data points have K-means-friendly structure in a two-dimensional domain (cf. the first subfigure of Fig.~\ref{fig: synthetic data}). This two-dimensional domain is a latent domain which we do not observe and we denote the latent representations of the data points as ${\bm h}_i$'s in this domain.
What we observe is ${\bm x}_i\in\mathbb{R}^{100}$ that is obtained via the following transformation:
\begin{align}\label{eq: sigsig}
{\bm x}_i = \sigma\left( {\bm U}\sigma({\bm W} {\bm h}_i)\right),
\end{align}
where ${\bm W}\in\mathbb{R}^{10\times 2}$ and ${\bm U}\in\mathbb{R}^{100\times 10}$ are matrices whose entries follow the zero-mean unit-variance i.i.d. Gaussian distribution, $\sigma(\cdot)$ is a sigmod function to introduce nonlinearity. Under the above generative model, recovering the K-means-friendly domain where ${\bm h}_i$'s live seems very challenging.

We generate four clusters, each of which has 2,500 samples and their geometric distribution on the 2-D plane is shown in the first subfigure of Fig.~\ref{fig: synthetic data} that we have seen before.
The other subfigures show the recovered 2-D data from ${\bm x}_i$'s using a number of
DR methods, namely, NMF \cite{lee1999learning}, local linear embedding (LLE) \cite{saul2003think}, Laplacian eigenmap (LapEig) \cite{ng2002spectral}  -- the first step of spectral clustering, and DEC \cite{xie2015unsupervised}.
We also present the result of using the formulation in \eqref{eq: plain} (DCN w/o reconstruction) which is a similar idea as in \cite{xie2015unsupervised}. For the three DNN-based methods (DCN, DEC, and SAE + KM), we use a four-layer forward network for dimensionality reduction,  where the layers have 100, 50, 10 and 2 neurons, respectively;
the reconstruction network used in DCN and SAE (and also in the per-training stage of DEC) is a mirrored version of the forward network.  
As one can see in Fig.~\ref{fig: synthetic data}, all the DR methods except the proposed DCN fail to map ${\bm x}_i$'s to a 2-D domain that is suitable for applying K-means.
In particular, DEC and DCN w/o reconstruction indeed give trivial solutions: the reduced-dimension data are separated to four clusters, and thus $\hat{L}$ is small.
But this solution is meaningless  since the data partitioning is arbitrary.

In the supplementary materials, two additional simulations with different generative model than \eqref{eq: sigsig} are presented, and similar results are observed. This further illustrates the DCN's ability of recovering clustering-friendly structure under different nonlinear generative models.

\vspace{-.25cm}
\subsection{Real-Data Validation}
\vspace{-.25cm}
In this section, we validate the proposed approach on several real-data sets which are all publicly available.

\vspace{-.25cm}
\subsubsection{Baseline methods}
\vspace{-.25cm}
We compare the proposed DCN with a variety of baseline methods:\\
\noindent
1) {\bf K-means (KM)}: The classic K-means \cite{lloyd1982least}.\\
\noindent
2) {\bf Spectral Clustering (SC)}: The classic SC algorithm \cite{ng2002spectral}.\\
\noindent
3) {\bf Sparse Subspace Clustering with Orthogonal Matching Pursuit (SSC-OMP)}  \cite{you2016scalable}: SSC is considered very competitive for clustering images; we use the newly proposed greedy version here for scalability.\\
\noindent
4) {\bf Locally Consistent Concept Factorization (LCCF)} \cite{cai2011locally}: LCCF is based on NMF with a graph Laplacian regularization and is considered state-of-the-art for document clustering.\\
\noindent
5) {\bf XRAY} \cite{kumar2013fast}: XRAY is an NMF-based document clustering algorithm that scales very well.\\
\noindent
6) {\bf NMF followed by K-means} (NMF+KM): This approach applies NMF for DR, and then applies K-means to the reduced-dimension data.\\
\noindent
7) {\bf Stacked Autoencoder followed by K-means (SAE+KM)}: This is also a two-stage approach. We use SAE for DR first and then apply K-means.\\
\noindent
8) {\bf Joint NMF and K-means (JNKM)} \cite{yang2016learning}: JNKM performs joint DR and $K$-means clustering as the proposed DCN does -- but the DR part is based on NMF.

9) {\bf Deep Embedded Clustering (DEC)} \cite{xie2015unsupervised}:  DEC performs joint DNN and clustering, where the loss function contains only clustering loss, without penalty on reconstruction as in our method. We use the code\footnote{https://github.com/piiswrong/dec} provided by the authors. For each experiment, we select the baselines that are considered most competitive and suitable for that application from the above pool.


\vspace{-.25cm}
\subsubsection{Evaluation metrics}
\vspace{-.25cm}
We adopt standard metrics for evaluating clustering performance. Specifically, we employ the following three metrics: normalized mutual information (NMI) \cite{cai2011locally}, adjusted Rand index (ARI) \cite{yeung2001details}, and clustering accuracy (ACC) \cite{cai2011locally}.
In a nutshell, all the above three measuring metrics are commonly used in the clustering literature, and all have pros and cons. But using them together suffices to demonstrate the effectiveness of the clustering algorithms. Note that NMI and ACC lie in the range of zero to one with one being the perfect clustering result and zero the worst. ARI is a value within $-1$ to $1$, with one being the best clustering performance and minus one the opposite.

\vspace{-.25cm}
\subsubsection{RCV1}
\vspace{-.25cm}

We first test the algorithms on a large-scale text corpus, namely, the Reuters Corpus Volume 1 Version 2 (RCV1-v2).
The RCV1-v2 corpus \cite{lewis2004rcv1} contains 804,414 documents, which were manually categorized into 103 different topics. We use a subset of the documents from the whole corpus. This subset contains 20 topics and $365,968$ documents and each document has a single topic label. As in \cite{srivastava2014dropout}, we pick the 2,000 most frequently used words (in the tf-idf form) as the features of the documents.

\begin{figure}
	\centering
	\includegraphics[width=.4\textwidth]{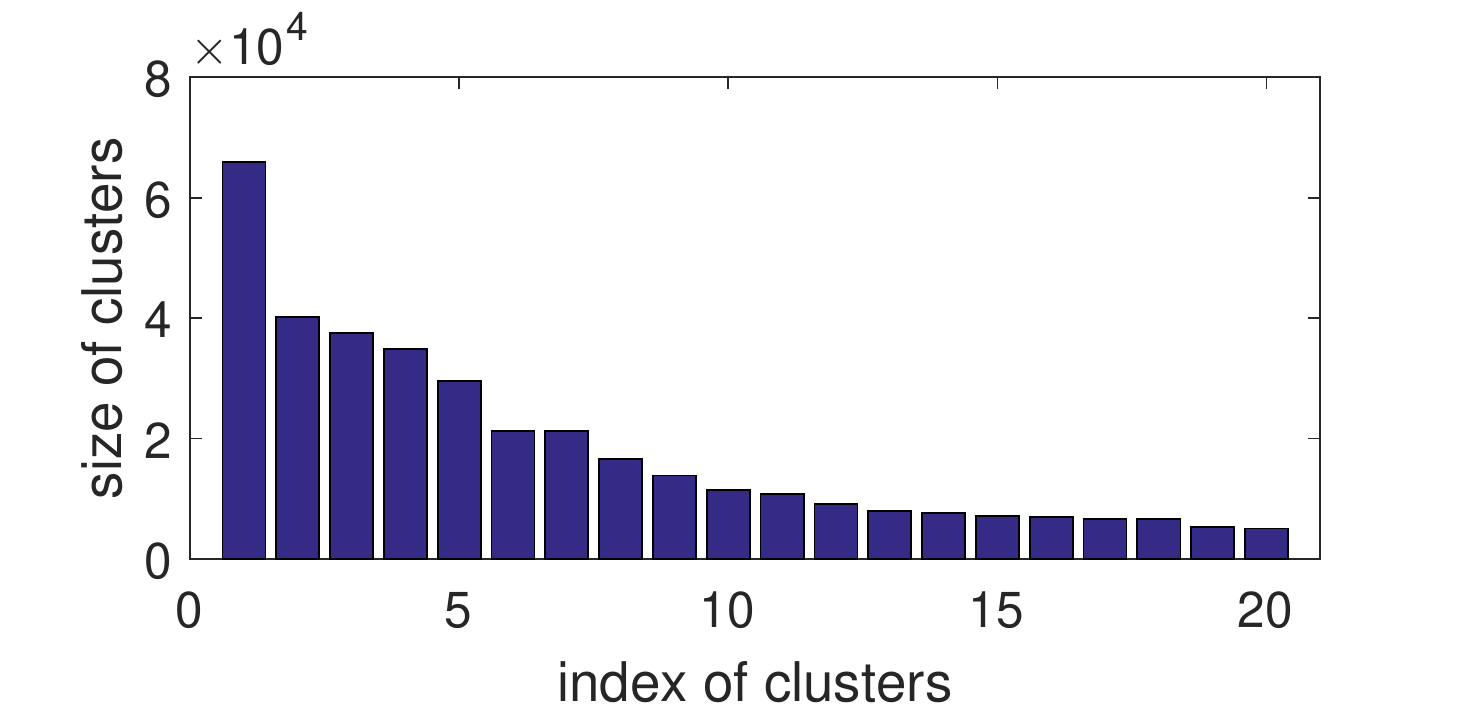}
	\caption{The sizes of 20 clusters in the experiment.}
	\vspace{-.5cm}
	\label{fig:RCV_size}
\end{figure}

We conduct experiments using different number of clusters.
Towards this end, we first sort the clusters according to the number of documents that they have in a descending order, and then apply the algorithms to the first 4, 8, 12, 16, 20 clusters, respectively.
Note that the first several clusters have many more documents compared to the other clusters (cf. Fig.~\ref{fig:RCV_size}).
This way, we gradually increase the number of documents in our experiments and create cases with much more unbalanced cluster sizes for testing the algorithms -- which means we gradually increase the difficulty of the experiments.
To avoid unrealistic tuning, for all the experiments, we use a DCN whose forward network has five hidden layers which have $2000, 1000, 1000, 1000, 50$ neurons, respectively. The reconstruction network has a mirrored structure.
We set $\lambda = 0.1$ for balancing the reconstruction error and the clustering regularization. 


Table \ref{Tab: rcv} shows the results given by the proposed DCN, SAE+KM, KM, and XRAY;
other baselines are not scalable enough to handle the RCV1-v2 dataset and thus are dropped.
One can see that for each case that we have tried, the proposed method gives clear improvement relative to the other methods. Particularly, the DCN approach outperforms the two-stage approach, i.e., SAE+KM, in almost all the cases and for all the evaluation metrics -- this clearly demonstrates the advantage of using the joint optimization criterion. We notice that the performance of DEC in this experiment is unsatisfactory, possibly because 1) this dataset is highly unbalanced (cf. Fig.~\ref{fig:RCV_size}), while DEC is designed to produce balanced clusters; 2) DEC gets trapped in trivial solutions, as we discussed in Sec~\ref{ch: background}.  

\begin{figure}[ht]
\centering
\begin{subfigure}[b]{0.23\textwidth}
	\includegraphics[width=\textwidth]{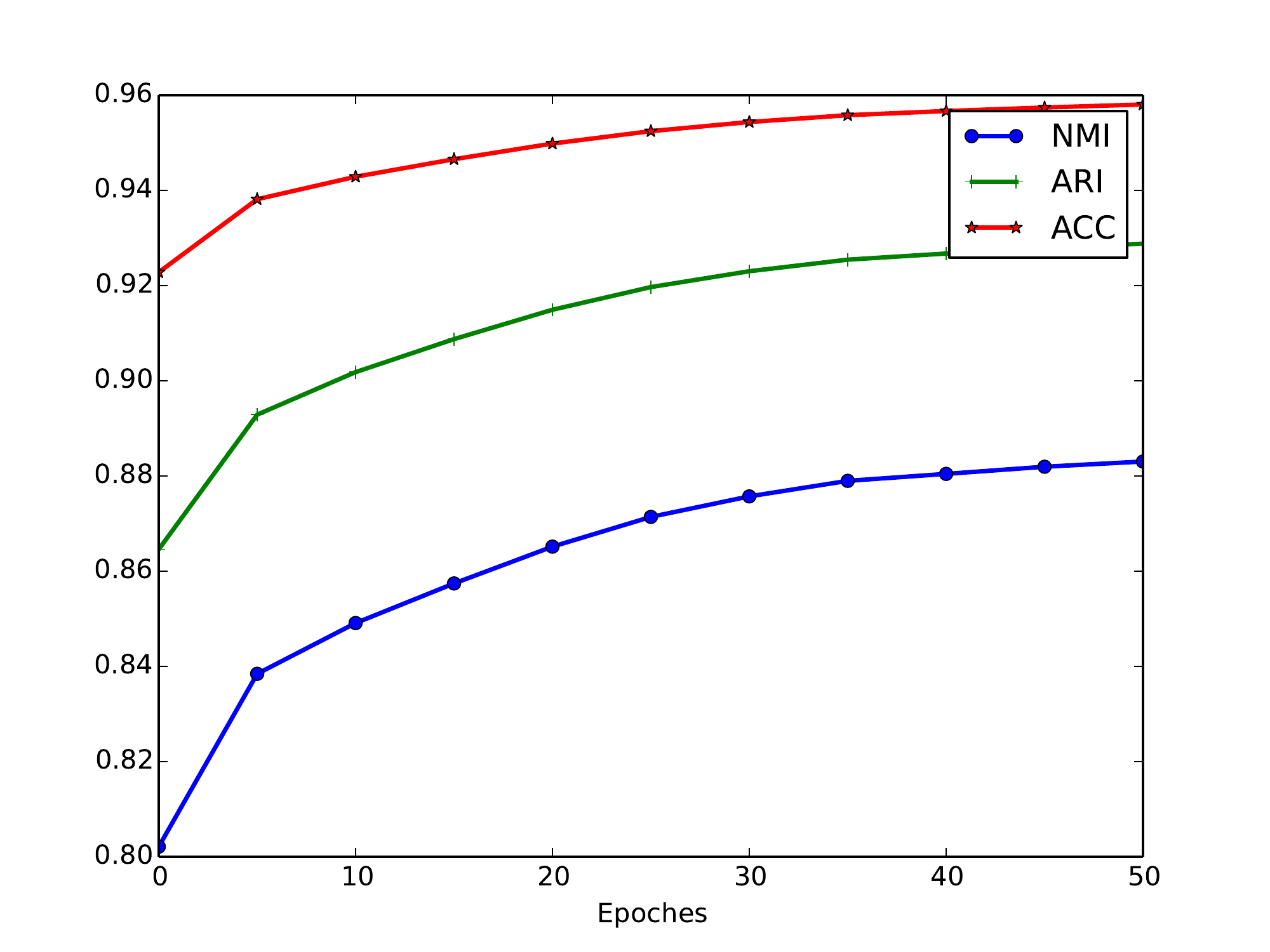}
	\vspace{0.3cm}
	\caption{Clustering performance metrics v.s. training epochs.}
	\label{fig:nmiariacc}
\end{subfigure}~
\begin{subfigure}[b]{0.24\textwidth}
	\includegraphics[width=\textwidth]{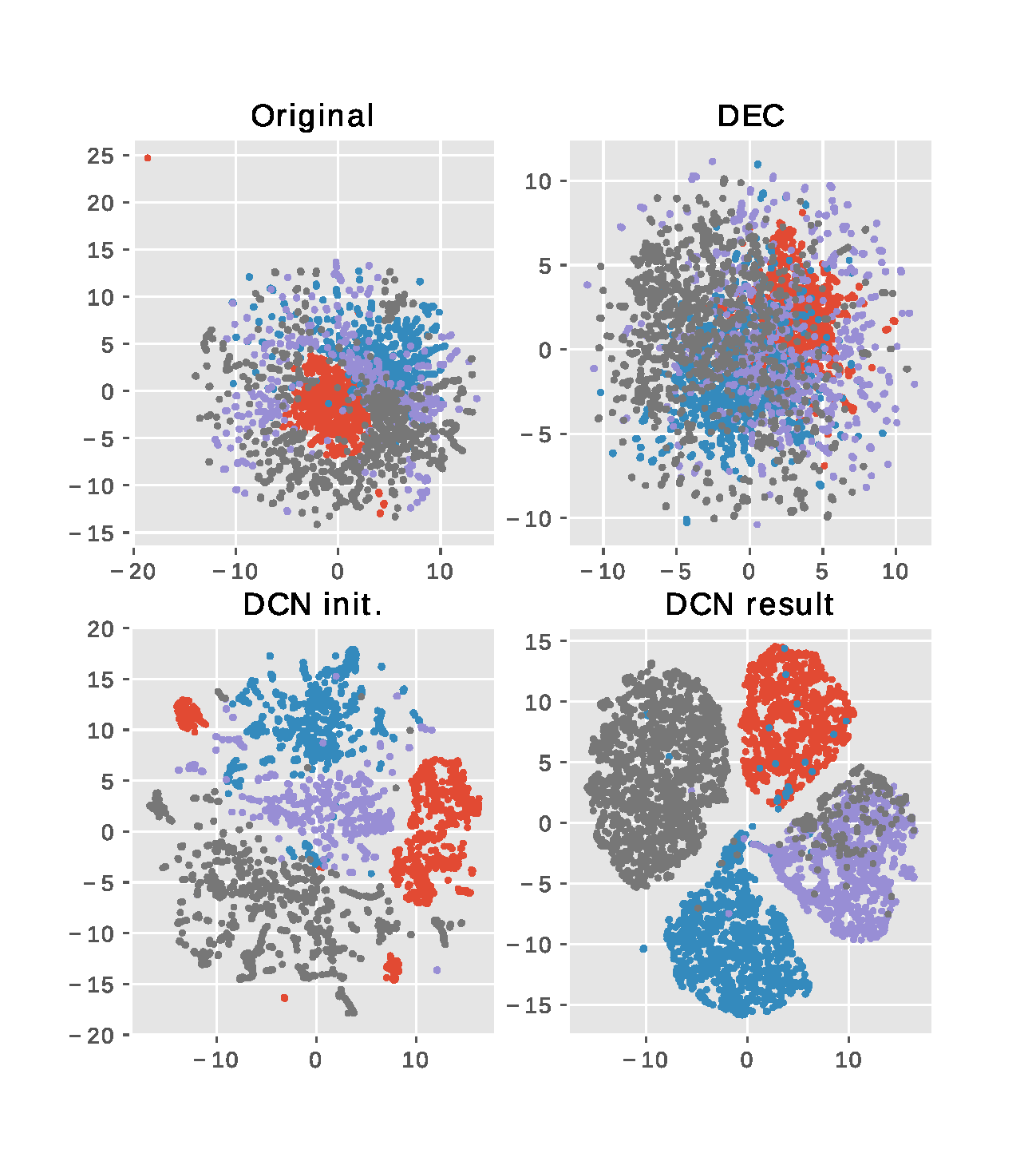}
	\caption{Visualization using t-SNE. From top-left to bottom-right: Original data, DEC result, DCN initialization, DCN result }
	\label{fig:dcn_dec}
\end{subfigure}
\caption{Visualization on the 4-clusters subset of RCV1-v2 }
\vspace{-0.5cm}
\end{figure}

Fig.~\ref{fig:nmiariacc} shows how NMI, ARI, and ACC change when the proposed algorithm runs from epoch to epoch.
One can see a clear ascending trend of every evaluation metric. This result shows that both the network structure and the optimization algorithm work towards a desired direction.
In the future, it would be intriguing to derive (sufficient) conditions for guaranteeing such improvement using the proposed algorithm.
Nevertheless, such empirical observation in Fig.~\ref{fig:nmiariacc} is already very interesting and encouraging.

\begin{table}[t]
	\centering
	\caption{Evaluation on the RCV1-v2 dataset}
	\vspace{0.2cm}
	\label{Tab: rcv}
	\resizebox{.75\linewidth}{!}{
	\begin{tabular}{lc|ccccc}
	\hline
		\multicolumn{2}{c|}{Methods} & DCN  & SAE+KM & KM &   DEC & XRAY \\
		\hline
		\multicolumn{1}{ c|  }{\multirow{3}{*}{4 Clust.} } & NMI & \textbf{0.76} & 0.73   & {0.62}  & 0.11 & 0.12 \\
		\multicolumn{1}{ c| }{} & ARI & \textbf{0.67} & 0.65  & 0.50  & 0.07 & -0.01 \\
		\multicolumn{1}{ c| }{} & ACC & \textbf{0.80} & 0.79  & 0.70  & 0.38 & 0.34 \\
		\hline
		\multicolumn{1}{ c|  }{\multirow{3}{*}{8 Clust.} } & NMI  & \textbf{0.63} & 0.60  & 0.57 & 0.10 & 0.24 \\
		\multicolumn{1}{ c| }{} & ARI & \textbf{0.46} & 0.42  & 0.38 & 0.05 & 0.09 \\
		\multicolumn{1}{ c| }{} & ACC & \textbf{0.63} & 0.62  & 0.59  & 0.24 & 0.39 \\
		\hline		
		\multicolumn{1}{ c|  }{\multirow{3}{*}{12 Clust.} } & NMI & \textbf{0.67} & 0.65  & 0.6 & 0.09  & 0.22 \\
		\multicolumn{1}{ c| }{} & ARI & \textbf{0.52} & 0.51  & 0.37 & 0.02 & 0.05 \\
		\multicolumn{1}{ c| }{} & ACC &  \textbf{0.60} & 0.56  & 0.54  & 0.18 & 0.29 \\
		\hline
		\multicolumn{1}{ c|  }{\multirow{3}{*}{16 Clust.} } & NMI & \textbf{0.62} & 0.60  & 0.56 & 0.09 & 0.23 \\
		\multicolumn{1}{ c| }{} & ARI & \textbf{0.36} & 0.35  & 0.30 & 0.02 & 0.04 \\
		\multicolumn{1}{ c| }{} & ACC & \textbf{0.51} & 0.50  & 0.48 & 0.17 & 0.29 \\
		\hline
		\multicolumn{1}{ c|  }{\multirow{3}{*}{20 Clust.} } & NMI & \textbf{0.61} & 0.59  & 0.58 & 0.08 & 0.25 \\
		\multicolumn{1}{ c| }{} & ARI & \textbf{0.33}  & \textbf{0.33} & 0.29 & 0.01 & 0.04 \\
		\multicolumn{1}{ c| }{} & ACC & \textbf{0.47} & 0.46  & \textbf{0.47} & 0.14  & 0.28 \\
		\hline
	\end{tabular}}
	\vspace{-0.5cm}
\end{table}
We visualize the 50-D learned embeddings of our network on the RCV1 4-clusters dataset, using t-SNE \cite{maaten2008visualizing}, as shown in Fig.~\ref{fig:dcn_dec}. We can see that the proposed DCN method learns much improved results compared to the initialization. Also, the DEC method does not get a desiable clustering result, possibly due to the imbalance clusters.

\vspace{-.25cm}
\subsubsection{20Newsgroup}
\vspace{-.25cm}
The 20Newsgroup corpus is a collection of 18,846 text documents which are partitioned into 20 different newsgroups. Using this corpus, we can observe how the proposed method works with a relatively small amount of samples. As the previous experiment, we use the tf-idf representation of the documents and pick the 2,000 most frequently used words as the features.
Since this dataset is small, we include more baselines that are not scalable enough for RCV1-v2.
Among them, both JNKM and LCCF are considered state-of-art for document clustering.
In this experiment, we use a DNN with three forward layers which have 250, 100, and 20 neurons, respectively.
This is a relatively `small network' since the 20Newsgroup corpus may not have sufficient samples to fit a large network. 
As before, the decoding network for reconstruction has a mirrored structure of the encoding part, and the baseline SAE+KM uses the same network for the autoencoder part.

Table~\ref{Tab:20News} summarizes the results of this experiment.
As one can see, LCCF indeed gives the best performance among the algorithms that do not use DNNs.
SAE+KM improves ARI and ACC quite substantially by involving DNN -- this suggests that the generative model may indeed be nonlinear. DCN performs even better by using the proposed joint DR and clustering criterion, which supports our motivation that a K-means regularization can help discover a clustering-friendly space.




\begin{table}[t]
	\centering
	\caption{Evaluation on the 20Newsgroup dataset.}
	\vspace{0.2cm}
	\label{Tab:20News}
	\resizebox{\linewidth}{!}{\huge
	\begin{tabular}{c|cccccccc}
		\hline
		Methods & DCN   & SAE+KM & LCCF  & NMF+KM & KM    & SC    & XARY  & JNKM \\
		\hline
		NMI   & \textbf{0.48} & 0.47  & 0.46  & 0.39  & 0.41  & 0.40  & 0.19  & 0.40 \\
		ARI   & \textbf{0.34} & 0.28  & 0.17  & 0.17  & 0.15  & 0.17  & 0.02  & 0.10 \\
		ACC   & \textbf{0.44} & 0.42  & 0.32  & 0.33  & 0.3   & 0.34  & 0.18  & 0.24 \\
		\hline
	\end{tabular}}%
		\vspace{-.5cm}
\end{table}%

\vspace{-.25cm}
\subsubsection{Raw MNIST}
\vspace{-.25cm}
In this and next subsections, we present two experiments using two versions of the MNIST dataset.
We first employ the raw MNIST dataset that has 70,000 data samples. Each sample is a $28\times 28$ gray-scale image containing a handwritten digit, i.e., one of $\{0,1,\ldots,9\}$. 
Same as \cite{xie2015unsupervised}, we use a  4-layers forward network and set the number of neurons to be 500, 500, 2000, and 10, respectively. The reconstruction network is still a `mirrored' version of the forward network. The hyperparameter $\lambda$ is set to $1$.
We use SSC-OMP, which is a scalable version of SSC, and KM as a baseline for this experiment.

\begin{table}[t]
	\centering
	\caption{Evaluation on the raw MNIST dataset.}
	\vspace{0.2cm}
	\label{Tab: MNIST_orig}
	\resizebox{.65\linewidth}{!}{\huge
	\begin{tabular}{l|ccccc}
		\hline
		Methods & DCN  & SAE+KM & DEC & KM & SSC-OMP \\
		\hline
		NMI & {\bf 0.81} & 0.73 & { 0.80} & 0.50  & 0.31\\
		ARI & {\bf 0.75} & 0.67 & {\bf 0.75} &0.37 & 0.13\\
		ACC & {0.83} & 0.80 & {\bf 0.84} &0.53 & 0.30\\
		\hline
	\end{tabular}}
	\vspace{-.5cm}
\end{table}

Table~\ref{Tab: MNIST_orig} shows results of applying DCN, SAE+KM, DEC, KM and SSC-OMP to the raw MNIST data -- the other baselines are not efficient enough to handle 70,000 samples and thus are left out. 
One can see that our result is on par with the result of DEC reported in \cite{xie2015unsupervised}, and both methods outperform other methods by a large margin. The DEC method performs very competitively on this dataset, possibly because it is designed to favor balanced clusters, which is the case for MNIST dataset. On the dataset RCV1-v2 with unbalanced clusters, the result of DEC is not as satisfactory, see Fig.~\ref{fig:dcn_dec}.  It is also interesting to note that our method yields approximately same results as DEC in this balanced case, but DCN also works well in unbalanced cases, as we have seen.


\vspace{-.25cm}
\subsubsection{Pre-Processed MNIST}
\vspace{-.25cm}
Besides the above experiment using the raw MNIST data, we also provide another interesting experiment using \emph{pre-processed} MNIST data.
The pre-processing is done by a recently introduced technique, namely, the scattering network (ScatNet) \cite{bruna2013invariant}. ScatNet is a cascade of multiple layers of wavelet transform, which is able to learn a good feature space for clustering / classification of images. Utilizing ScatNet, the work in \cite{you2016scalable} reported very promising clustering results on MNIST using SSC-OMP. Our objective here is to see if the proposed DCN can further improve the performance from SSC-OMP. Our idea is simple: SSC-OMP is essentially a procedure of constructing a similarity matrix of the data; after obtaining this matrix, it performs K-means on the rows of a matrix comprising several selected eigenvectors of the similarity matrix \cite{ng2002spectral}. Therefore, it makes sense to treat the whole ScatNet + SSC-OMP procedure as pre-processing for performing K-means, and one can replace K-means by DCN to improve performance.

The results are shown in Table \ref{Tab: MNIST}. One can see that the proposed method exhibits the best performance among the algorithms. We note that the result of using KM on the data processed by ScatNet and SSC-OMP is worse than that was reported in \cite{you2016scalable}. This is possibly because we use all the 70,000 samples, while only a subset was selected for conducting the experiments in \cite{you2016scalable}.

This experiment is particularly interesting since it suggests that for any clustering algorithm that employs K-means as a key component, e.g., spectral clustering and sparse subspace clustering, one can use the proposed DCN to replace K-means and a better result can be expected. This is meaningful since many datasets are originally not suitable for K-means due to the nature of the data -- but after pre-processing (e.g., kernelization and eigendecomposition), the pre-processed data is already more K-means-friendly, and using the proposed DCN at this point can further strengthen the result.

\begin{table}
	\centering
	\caption{Evaluation on pre-processed MNIST}
	\vspace{0.2cm}
   \resizebox{.65\linewidth}{!}{\huge
	\label{Tab: MNIST}
	\begin{tabular}{l|ccc}
		\hline
		 Methods & DCN  & SAE+KM &  KM (SSC-OMP)  \\
		\hline
		 NMI & {\bf 0.88} & 0.86 &  {0.85}\\		
		 ARI & {\bf 0.89} & 0.86 &  0.82\\
		 ACC & {\bf 0.95} & 0.93 &  0.86\\
		\hline
	\end{tabular}}
	\vspace{-.3cm}
\end{table}

\vspace{-.25cm}
\subsubsection{Parameter Selection}
\begin{figure}
	\centering
	\includegraphics[width=.25\textwidth]{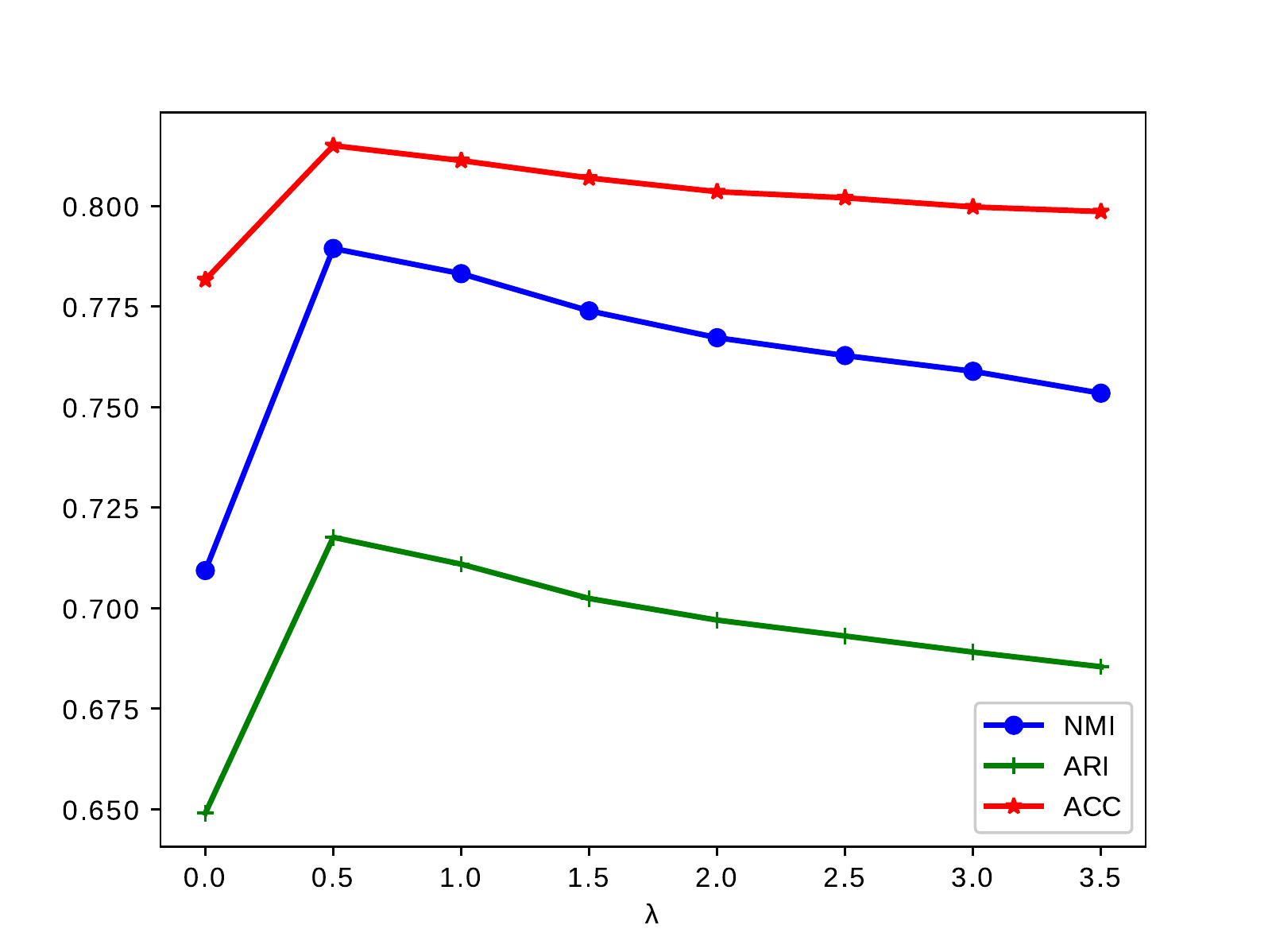}
	\vspace{-.5cm}
	\caption{Clustering performance on MNIST with different $\lambda$.}
	\vspace{-.5cm}
	\label{fig:params}
\end{figure}

\vspace{-.25cm}
The parameter $\lambda$ is important, since it trades off between the reconstruction objective and the clustering objective.
As we see from the experiments, the proposed DCN works well with an appropriately chosen $\lambda$. Moreover, our experience suggests that the performance of our approach is insensitive to the exact value of $\lambda$. Fig. \ref{fig:params} shows how the proposed method performs with different $\lambda$ on the MNIST dataset. As we can see, although there is degradation of performance as $\lambda$ gets inappropriately large, the degradation is mild.  The proposed method gives satisfactory result for a range of $\lambda$.

\vspace{-.25cm}
\section{Conclusion}
\vspace{-.25cm}
In this work, we proposed a joint DR and K-means clustering approach where the DR part is accomplished via learning a deep neural network. Our goal is to automatically map high-dimensional data to a latent space where K-means is a suitable tool for clustering. We carefully designed the network structure to avoid trivial and meaningless solutions and proposed an effective and scalable optimization procedure to handle the formulated challenging problem. Synthetic and real data experiments showed that the algorithm is very effective on a variety of datasets.

\section*{Acknowledgements}
This work is supported by National Science Foundation under Projects NSF IIS-1447788, NSF ECCS-1608961,
and NSF CCF-1526078. The GPU used in this work was kindly donated by NVIDIA.

\bibliography{ref}
\bibliographystyle{icml2017}
\newpage
\setcounter{section}{0}
\setcounter{figure}{0}
\setcounter{table}{0}
{\Large Supplementary material for ``Towards K-means-friendly Spaces: Simultaneous Deep Learning and Clustering"}
\section{Additional Synthetic-Data Experiments}
\subsection{Additional Generative Models}
\begin{figure}[ht]
	\centering
	\includegraphics[width=0.45\textwidth]{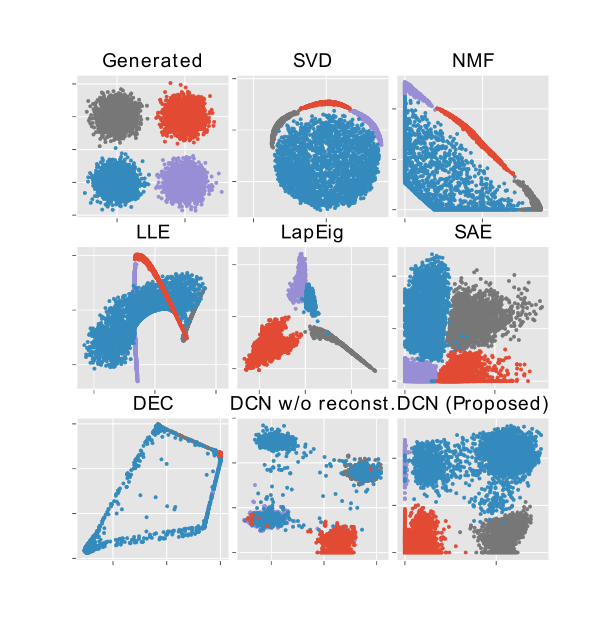}
	\caption{The generated latent representations $\{{\bm h}_i\}$ in the 2-D space and the recovered 2-D representations from ${\bm x}_i\in\mathbb{R}^{100}$, where ${\bm x}_i =\left( \sigma({\bm W} {\bm h}_i)\right)^2$.}
	\label{fig:9methods_2}
\end{figure}

In this section, we provide two more examples to illustrate the ability of DCN in recovering K-means-friendly spaces under different generative models.
We first consider the transformation as follows:
\begin{equation}\label{eq:nonlinear_1}
	{\bm x}_i = \left( \sigma({\bm W} {\bm h}_i)\right)^2,
\end{equation}
where $\sigma(\cdot)$ is the {sigmoid} function as before and ${\bm W} \in \R^{100\times2}$ is similarly generated as in the paper.  We perform elementwise squaring on the result features to further complicate the generating process.
The corresponding results can be seen in Fig.~\ref{fig:9methods_2} of this supplementary document.
One can see that a similar pattern as we have observed in the main text is also presented here: The proposed DCN recovers a 2-D K-means-friendly space very well and the other methods all fail.

\begin{figure}[ht]
	\centering
	\includegraphics[width=0.45\textwidth]{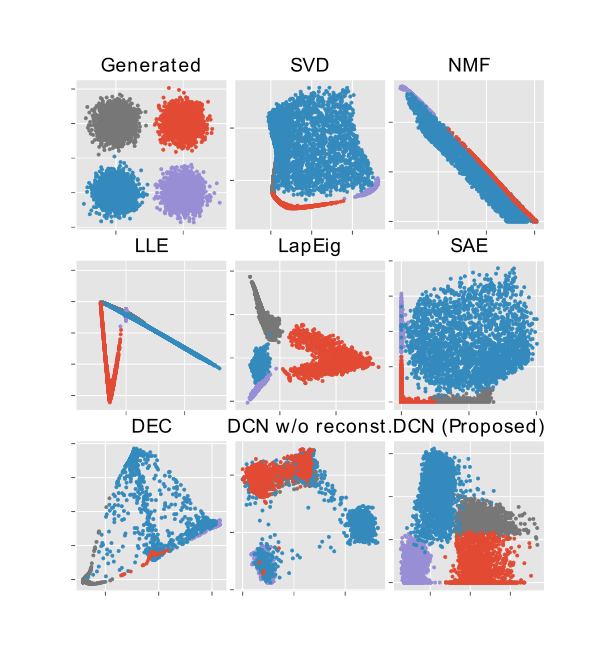}
	\caption{The generated latent representations $\{{\bm h}_i\}$ in the 2-D space of the recovered 2-D representations from ${\bm x}_i\in\mathbb{R}^{100}$, where ${\bm x}_i =  \text{tanh}\left( \sigma({\bm W} {\bm h}_i)\right)$.}
	\label{fig:9methods_3}
\end{figure}

In Fig.~\ref{fig:9methods_3}, we test the algorithms under the generative model
\begin{equation}\label{eq:nonlinear_2}
	{\bm x}_i =  \text{tanh}\left( \sigma({\bm W} {\bm h}_i)\right),
\end{equation}
where ${\bm W} \in \mathbb{R}^{100\times 2}$. Same as before, the proposed DCN gives very clear clusters in the recovered 2-D space.

The results in this section and the synthetic-data experiment presented in main text are encouraging: Under a variety of complicated nonlinear generative models, DCN can output clustering-friendly latent representations.

\section{Additional Real-Data Experiments}
\subsection{Pendigits}
Beside the real datasets in the paper, we also conduct experiment on the Pendigits dataset.
The Pendigits dataset consists of 10,992 data samples. Each sample records 8 coordinates on a tablet, on which a subject is instructed to write the digits from $0$ to $9$. So each sample corresponds to a vector of length 16, and represents one of the digits. Note that this dataset is quite different from MNIST -- each digit in MNIST is represented by an image (pixel values) while digits in Pendigits are represented by 8 coordinates of the stylus when a person was writing a certain digit.
Since each digit is represented by a very small-size vector of length 16,
we use a small network who has three forward layers which are with 16, 16, and 10 neurons.
Table~\ref{Tab:Pendigits} shows the results: The proposed methods give the best clustering performance compared to
the competing methods, and the methods using DNNs outperform the `shallow' ones that do not use neural networks for DR.

\begin{table}[t]
	\centering
	\caption{Evaluation on the Pendigits dataset}
	\label{Tab:Pendigits}
	\resizebox{.65\linewidth}{!}{
	\begin{tabular}{c|cccc}
		\hline
		Methods & {DCN} & SAE+KM & SC    & KM  \\
		\hline
		NMI   & \textbf{0.69} & 0.65  & 0.67  & 0.67  \\
		ARI   & \textbf{0.56} & 0.53  & 0.55  & 0.55  \\
		ACC   & \textbf{0.72} & 0.70  & 0.71  & 0.69  \\
		\hline
	\end{tabular}%
	\label{tab:addlabel}}%
	\vspace{-.5cm}
\end{table}%
\vspace{-.15cm}

\subsection{DCN as Feature Learner}
We motivate and develop DCN as a clustering method that directly works on unlabeled data. 
In practice, DCN can also be utilized as a feature-learning method when training samples are available -- i.e., one can feed labeled training data to DCN, tune the parameters of the network
to learn well clustered latent representations of the training samples, and then use the trained DCN (to be specific, the forward network) to reduce dimension of unseen testing data.

Here, we provide some additional results to showcase the feature-learning ability of DCN. We perform a 5-fold cross-validation experiment on the raw MNIST dataset, where each fold is a 80/20 training/testing random split. The performance of SAE+KM on the training sets is presented as a baseline.
\begin{table}
	\caption{The mean (stand deviation) of the evaluation results of the 5-fold cross-validation on MNIST.}
	\resizebox{\linewidth}{!}{\begin{tabular}{lccc}
		\hline
		& NMI & ARI & ACC \\
		\hline
		DCN-Training & 0.80 (0.001) &  0.74 (0.002) & 0.83 (0.002) \\
		DCN-Testing & 0.81 (0.003) & 0.75 (0.005) & 0.83 (0.004) \\
		SAE+KM & 0.73 (0.001) & 0.67 (0.002) & 0.80 (0.002) \\
		\hline
	\end{tabular}}
	\label{tab: dcn-testing}
\end{table}

The obtained NMI, ARI, and ACC (mean and standard deviation) are listed in Table. \ref{tab: dcn-testing}.
One can see that the training and testing stages of DCN output similar results, which is rather encouraging.
This experiment suggests that DCN is very promising as a representation learner. 

\section{Detailed Settings of Real-Data Experiments}
\subsection{Algorithm Parameters}
There is a set of parameters in the proposed algorithm which need to be pre-defined.
Specifically, the learning rate $\alpha$, the number of epochs $T$ (recall that one epoch responds to a pass of all the data samples through the network), and the balancing regularization parameter $\lambda$. These parameters vary from case to case since they are related to a number of factors, e.g., dimension of the data samples, total number of samples, scale (or energy) of the samples, etc. In practice, a reasonable way to tune these parameters is through observing the performance of the algorithm under various parameters on a small validation subset whose labels are known.

Note that the proposed algorithm has two stages, i.e., pre-training and the main algorithm and they usually use two different sets of parameters since the algorithmic structure of the two stages are quite different (to be more precise, the pre-training state does not work with the whole network but only deals with a pair of encoding-decoding layers greedily). Therefore, we distinguish the parameters of the two stages as listed in Table~\ref{Tab: notations}, to better describe the settings. 

We implement SGD for solving the subproblem w.r.t. ${\cal X}$ using the Nesterov-type acceleration \cite{nesterov2013introductory}, the mini-batch version, and the momentum method. Batch normalization \cite{ioffe2015batch} that is recently proven to be very effective for training supervised deep networks is also employed.
Through out the experiments, the momentum parameter is set to be $0.9$,
the mini-batch size is selected to be $\approx 0.01\times N$,
and the other parameters are adjusted accordingly in each experiments -- which will be described in detail in the next section.
\subsection{Network Parameters}
The considered network has two parts, namely, the forward encoding network that reduces the dimensionality of the data and
the decoding network that reconstructs the data.
We let two networks to have a mirrored structure of each other.
There are also two parameters of a forward network, i.e., the width of each layer (number of neurons) and the depth of the network (number of layers). There is no strict rule for setting up these two parameters, but the rule of thumb is to adjust them according the amounts of data samples of the datasets and the dimension of each sample. Using a deeper and wider network may be able to better capture the underlying nonlinear transformation of the data, as the network has more degrees of freedom. However, finding a large number of parameters accurately requires a large amount of data since the procedure can be essentially considered as solving a large system of nonlinear equations -- and finding more unknowns needs more equalities in the system, or, data samples in this case. Therefore, there is a clear trade-off between network depth/width and the overall performance. 

\begin{table}[t]
	\centering
	\caption{List of parameters used in DCN.}
	\begin{tabular}{r|l}
		\hline
		Notations & Meaning\\
		\hline
		$\lambda$ & regularization parameter \\
		$\alpha_p$ & base pre-training stepsize \\
		$\alpha_l$ & base learning stepsize \\
		$T_p$    & pre-traing epochs \\
		$T_l$    & learning epochs\\
		\hline
	\end{tabular}
	\label{Tab: notations}
\end{table}

\subsection{Detailed Parameter Settings}
The detailed parameter settings for experiments on RCV1-v2 are shown in Tables~\ref{Tab: rcv1_1}. Parameter settings for 20Newsgroup, raw MNIST, pre-processed MNIST, and Pendigits are shown in Tables~\ref{Tab: 20News},~\ref{Tab: rawMNIST},~\ref{Tab: ProcMNIST}, and \ref{Tab: pendigit}, respectively.

\section{More Discussions}
\begin{figure}[ht]
	\centering
	\includegraphics[width=0.45\textwidth]{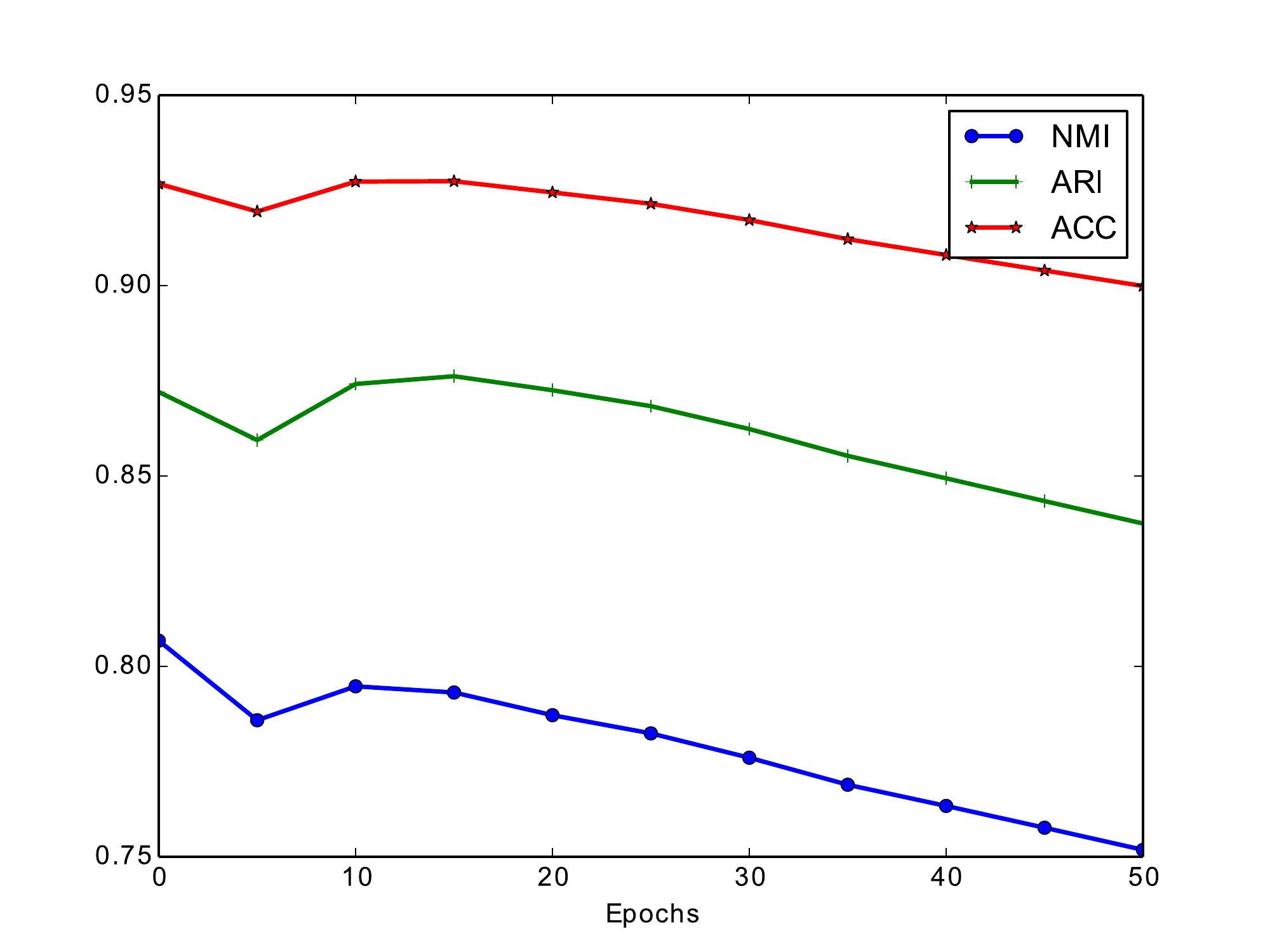}
	\caption{Clustering performance degrades when training with only reconstruction error term. This is in sharp contrast with Figure 5(a) in the paper, where clustering performance improves when training the proposed DCN model.}
	\label{fig:sae}
\end{figure}

We have the following several more points as further discussion:

\begin{enumerate}
	\item We have observed that runing SAE for epochs may even worsen the clustering performance in the two-stage approach. In Fig. \ref{fig:sae}, we show how the clustering performance indexes change with the epochs when we run SAE without K-means regularization.One can see that the performance in fact becomes worse compared to merely using pre-training (i.e., initialization).
	This means that using the SAE does not necessarily help clustering -- and this supports our motivation for adding a K-means-friendly structure-enhancing regularization.
	
	\item To alleviate the effect brought by the intrinsic randomness of the algorithms, e.g., random initialization of pre-training, the reported results are all obtained via running the experiments several times and taking average (specifically, we run the experiments with smaller size, i.e., 20Newsgroup, raw and processed MNIST, and Pendigits for ten times and the results of the much larger dataset RCV-v2 are average of five runs; the results for DEC in Table 1 is from a single run.). 
	Therefore, the presented results reflect the performance of the algorithms in an average sense.
	

	\item We treat this work as a proof-of-concept: Joint DNN learning and clustering is a highly viable task according to our design and experiments. In the future, many practical issues will be investigated -- e.g., designing theory-backed ways of setting up network and algorithm parameters. 
	Another very intriguing direction is of course to design convergence-guaranteed algorithms for optimizing the proposed criterion and its variants. We leave these interesting considerations for future work.
\end{enumerate}

\begin{table}[ht]
	\centering	
	\caption{Parameter settings for RCV1-v2}
	\label{Tab: rcv1_1}
	\begin{tabular}{r|r}
		\hline
		\multicolumn{1}{c|}{parameters} & \multicolumn{1}{c}{description} \\
		\hline
		${\bm f}({\bm x}_i; {\cal W})$: $\mathbb{R}^M\rightarrow \mathbb{R}^R$   & \multicolumn{1}{l}{$M=2,000$ and $R=50$} \\
		Sample size $N$ & \multicolumn{1}{l}{178,603 or 267,466}\\
		forward net. depth & \multicolumn{1}{l}{5 layers } \\
		layer width & \multicolumn{1}{l}{2000/1000/1000/1000/50} \\
		$\lambda$ & \multicolumn{1}{l}{0.1} \\
		$\alpha_p$ & \multicolumn{1}{l}{0.01} \\
		$\alpha_l$ & \multicolumn{1}{l}{0.05} \\
		$T_p$    & \multicolumn{1}{l}{50} \\
		$T_l$    & \multicolumn{1}{l}{50} \\
		\hline
	\end{tabular}%
\end{table}
\begin{table}[ht]
	\centering	
	\caption{Parameter settings for 20Newsgroup}
	\label{Tab: 20News}
	\begin{tabular}{r|r}
		\hline
		\multicolumn{1}{c|}{parameters} & \multicolumn{1}{c}{description} \\
		\hline
		${\bm f}({\bm x}_i; {\cal W})$: $\mathbb{R}^M\rightarrow \mathbb{R}^R$   & \multicolumn{1}{l}{$M=2,000$ and $R=20$} \\
		Sample size $N$ & \multicolumn{1}{l}{18,846}\\
		forward net. depth & \multicolumn{1}{l}{3 layers } \\
		layer width & \multicolumn{1}{l}{250/100/20} \\
		$\lambda$ & \multicolumn{1}{l}{10} \\
		$\alpha_p$ & \multicolumn{1}{l}{0.01} \\
		$\alpha_l$ & \multicolumn{1}{l}{0.001} \\
		$T_p$    & \multicolumn{1}{l}{10} \\
		$T_l$    & \multicolumn{1}{l}{50} \\
		\hline
	\end{tabular}%
\end{table}
\begin{table}[ht]
	\centering	
	\caption{Parameter settings for raw MNIST}
	\label{Tab: rawMNIST}
	\begin{tabular}{r|r}
		\hline
		\multicolumn{1}{c|}{parameters} & \multicolumn{1}{c}{description} \\
		\hline
		${\bm f}({\bm x}_i; {\cal W})$: $\mathbb{R}^M\rightarrow \mathbb{R}^R$   & \multicolumn{1}{l}{$M=784$ and $R=50$} \\
		Sample size $N$ & \multicolumn{1}{l}{70,000}\\
		forward net. depth & \multicolumn{1}{l}{4 layers } \\
		layer width & \multicolumn{1}{l}{500/ 500/ 2000/10} \\
		$\lambda$ & \multicolumn{1}{l}{0.05} \\
		$\alpha_p$ & \multicolumn{1}{l}{0.01} \\
		$\alpha_l$ & \multicolumn{1}{l}{0.05} \\
		$T_p$    & \multicolumn{1}{l}{50} \\
		$T_l$    & \multicolumn{1}{l}{50} \\
		\hline
	\end{tabular}%
\end{table}
\begin{table}[ht]
	\centering	
	\caption{Parameter settings for Pre-Processed MNIST}
	\label{Tab: ProcMNIST}
	\begin{tabular}{r|r}
		\hline
		\multicolumn{1}{c|}{parameters} & \multicolumn{1}{c}{description} \\
		\hline
		${\bm f}({\bm x}_i; {\cal W})$: $\mathbb{R}^M\rightarrow \mathbb{R}^R$   & \multicolumn{1}{l}{$M=10$ and $R=5$} \\
		Sample size $N$ & \multicolumn{1}{l}{70,000}\\
		forward net. depth & \multicolumn{1}{l}{3 layers } \\
		layer width & \multicolumn{1}{l}{50/ 20/ 5} \\
		$\lambda$ & \multicolumn{1}{l}{0.1} \\
		$\alpha_p$ & \multicolumn{1}{l}{0.01} \\
		$\alpha_l$ & \multicolumn{1}{l}{0.01} \\
		$T_p$    & \multicolumn{1}{l}{10} \\
		$T_l$    & \multicolumn{1}{l}{50} \\
		\hline
	\end{tabular}%
\end{table}

\begin{table}[ht]
	\centering	
	\caption{Parameter settings for Pendigits}
	\label{Tab: pendigit}
	\begin{tabular}{r|r}
		\hline
		\multicolumn{1}{c|}{parameters} & \multicolumn{1}{c}{description} \\
		\hline
		${\bm f}({\bm x}_i; {\cal W})$: $\mathbb{R}^M\rightarrow \mathbb{R}^R$   & \multicolumn{1}{l}{$M=16$ and $R=10$} \\
		Sample size $N$ & \multicolumn{1}{l}{10,992}\\
		forward net. depth & \multicolumn{1}{l}{3 layers } \\
		layer width & \multicolumn{1}{l}{50/ 16/ 10} \\
		$\lambda$ & \multicolumn{1}{l}{0.5} \\
		$\alpha_p$ & \multicolumn{1}{l}{0.01} \\
		$\alpha_l$ & \multicolumn{1}{l}{0.01} \\
		$T_p$    & \multicolumn{1}{l}{50} \\
		$T_l$    & \multicolumn{1}{l}{50} \\
		\hline
	\end{tabular}%
\end{table}

\end{document}